# An Axiomatic Framework for Belief Updates

David E. Heckerman

Medical Computer Science Group
Knowledge Systems Laboratory
Stanford University
Medical School Office Building, Room 215
Stanford, California 94305

## Abstract

In the 1940's, a physicist named Cox provided the first formal justification for the axioms of probability based on the subjective or Bayesian interpretation. He showed that if a measure of belief satisfies several fundamental properties, then *the measure* must be some monotonic transformation of a probability. In this paper, measures of *change* in belief or *belief updates* are examined. In the spirit of Cox, properties for a measure of change in belief are enumerated. It is shown that if a measure satisfies these properties, it must satisfy other restrictive conditions. For example, it is shown that belief updates in a probabilistic context must be equal to some monotonic transformation of a likelihood ratio. It is hoped that this formal explication of the belief update paradigm will facilitate critical discussion and useful extensions of the approach.

## Introduction

As researchers in artificial intelligence have begun to tackle real-world domains such as medical diagnosis, mineral exploration, and financial planning, there has been increasing interest in the development and refinement of methods for reasoning with uncertainty. Much of the work in this area has been focused on methods for the representation and manipulation of measures of *absolute* belief, quantities which reflect the absolute degree to which propositions are believed. There has also been much interest in methodologies which focus on measures of *change* in belief or *belief updates*[1], quantities which reflect the degree to which beliefs in propositions change when evidence about them becomes known. Such methodologies include the MYCIN certainty factor model [5], the PROSPECTOR scoring scheme [6], and the application of Dempster's Rule to the combination of "weights of evidence" [7].

In this paper, a formal explication of the belief update paradigm is given. The presentation is modeled after the work of a physicist named R.T. Cox. In 1946, Cox [8] enumerated a small set of intuitive properties for a measure of *absolute* belief and proved that *any* measure that satisfies these properties must be some monotonic transformation of a probability. In the same spirit, a set of properties or axioms that are intended to capture the notion of a belief update are enumerated. It is then shown that these properties place strong restrictions on measures of change in belief. For example, it is shown that the only measures which satisfy the properties in a probabilistic context are monotonic transformations of the likelihood ratio $\lambda(H,E,e) = p(E|H,e)/p(E|\neg H,e)$, where H is a hypothesis, E is a piece of evidence relevant to the hypothesis, and e is background information.

It should be emphasized that the goal of this axiomization is not to prove that belief updates can only take the form described above. Rather, it is hoped that a formal explication of the update paradigm will stimulate constructive research in this area. For example, the axioms presented here can serve as a tool for the identification and communication of dissatisfaction with the update approach. Given the properties for a belief update, a researcher may be able to pinpoint the source of his dissatisfaction and criticize one or more of the properties directly. In addition, a precise characterization of the update paradigm can be useful in promoting consistent use of the approach. This is important as methodologies which manipulate measures of change in belief have been used inconsistently in the past [9]. Finally, it is hoped that the identification of assumptions underlying the paradigm will allow implementors to better judge the appropriateness of the method for application in a given domain.

Although there has been much discussion concerning the foundations of methodologies which focus on measures of absolute belief [8, 10, 11], there have been few efforts directed at measures of change in belief. Notable exceptions are the works of Popper [12] and Good [13]. Popper proposed a set of properties or axioms that reflect his notion of belief update which he called *corroboration* and Good showed that the likelihood ratio $\lambda(H,E,e)$ satisfies these properties [13]. Unfortunately, Popper's desiderata are somewhat non-intuitive and restricted to a probabilistic context. The axiomization here is offered as an alternative.

## Scope of the axiomization

The process of reasoning under uncertainty can be decomposed into three components: problem formulation, belief assignment, and belief entailment. *Problem formulation* refers to the process of enumerating the propositions or events of interest as well as the possible outcomes of each proposition. *Belief assignment* refers to the process of constructing and measuring beliefs about propositions of interest. Finally, *belief entailment* refers to the process of deriving beliefs from beliefs assessed in the second phase.

It must be emphasized that most methods for reasoning with uncertainty, including those in which belief updates are central, focus primarily on the third component described above.[2] Indeed, it could be argued that a significant portion of the controversy over the adequacy of various methods for reasoning with uncertainty has stemmed from a lack of appreciation of this fact.[3] The axiomization of the belief update paradigm presented here similarly restricts its focus to the process of belief entailment.

## Fundamental properties for a measure of absolute belief

Before presenting the axiomization for belief updates, it is useful to consider the properties Cox enumerated for a measure of absolute belief. This discussion will help motivate the characterization of measures of change in belief as it is similar in spirit. In addition, several of the properties for a measure of absolute belief will be needed for the explication of the belief update paradigm.

The first property proposed by Cox concerns the nature of propositions to which beliefs can be assigned. He asserted that propositions must be defined precisely enough so that it would be possible to determine whether a proposition is indeed true or false. That is, a proposition should be defined clearly enough that an all-knowing *clairvoyant* could determine its truth or falsehood. This requirement will be called the *clarity* property.[4]

A second property asserted by Cox is that it is possible to assign a degree of belief to any proposition which is precisely defined. This property will be termed the *completeness* property.

Cox also asserted that a measure of belief can vary continuously between values of absolute truth and falsehood and that the continuum of belief can be represented by a single real number. For definiteness, it will be assumed that larger numbers correspond to larger degrees of belief. The use of a single real number to represent continuous measures of belief will be called the *scalar continuity* property.

Another fundamental assumption made by Cox is that the degree of belief for a proposition will depend on the current state of information of the individual assessing the belief. To emphasize this, the term P|e, read "P given e," will be used to denote the degree of belief in proposition P for some individual with information e. This assumption will be termed the *context dependency* property.

Now consider two propositions P and Q. If P and Q are logically equivalent then P|e = Q|e. That is, if P is true only when Q is true and vice-versa, an individual should believe each of them with equal conviction. Thus, for example, it must be that XY|e = YX|e where XY denotes the proposition "X AND Y." This axiom will be called the *consistency* property.

Another property asserted by Cox is that the belief in the conjunction PQ should be related to the belief in P alone and to the belief in Q given that P is true. Formally, Cox proposed that there exists some function F such that

$$PQ|e = F(P|e, Q|Pe). \qquad (1)$$

The function is asserted to be continuous in both arguments and monotonically increasing[5] in each argument when the other is held constant. This property captures the notion that individuals commonly assign belief to events conditioned on the truth of another. This property will be termed the *hypothetical conditioning* property.

Finally, Cox asserted that the belief in ¬P (not P) should be related to the belief in P. Formally, he asserted that there should be some function G such that

$$\neg P|E = G(P|e). \qquad (2)$$

The only restrictions placed on G are that it be continuous and monotonically decreasing. This assumption will be called the *complementarity* property.

After enumerating these properties, Cox proved that *any* measure which satisfies them must also satisfy the relations:

$$0 \leq H(P|e) \leq 1 \qquad (3)$$

$$H(TRUE|e) = 1 \qquad (4)$$

$$H(PQ|e) = H(P|e) \cdot H(Q|Pe) \quad \text{(product rule)} \qquad (5)$$

$$H(P|e) + H(\neg P|e) = 1. \quad \text{(sum rule)} \qquad (6)$$

where H is a monotonically increasing function. However, (3) - (6) implies that H(P|e) is a probability. That is, (3) - (6) correspond to the axioms of probability theory. Therefore, Cox demonstrated that if one accepts the above properties, one must accept that probability is the only admissible measure of absolute belief.

Cox's proof is simple and elegant. The reader is urged to consult the original work to gain a better appreciation of the argument. The work also contains an interesting discussion by Cox arguing for each of the properties he describes.

In the sections to follow, an argument analogous to Cox's for belief updates is presented. As mentioned above, there will be little effort made to justify the properties enumerated. Instead, it is hoped that this exposition will foster constructive discussion about the usefulness of the update paradigm.

## Fundamental properties for a measure of change in belief

Suppose an individual with background information e has a belief in some hypothesis H for which a piece of evidence E becomes known. The basic assumption of the update paradigm is that a belief update, denoted U(H,E,e), in conjunction with the *prior* belief, H|e, is sufficient for determining the *posterior* belief H|Ee. More formally, it is assumed that there exists some function f such that

$$H|Ee = f(U(H,E,e), H|e). \qquad (7)$$

In the paradigm, the quantities U(H,E,e), H|e, and H|Ee are all single real numbers. In addition, it is required that the function f be continuous in both arguments and that f be monotonically increasing in each argument when the other is held constant.

Equation (7) is the definition of a belief update. Note that only the context dependency property and the scalar continuity property for a measure of absolute belief have been assumed in this definition.

It is useful to view the function f in (7) as an *updating procedure* which operates on a prior belief and returns a posterior belief. The procedure, in turn, is parameterized by the single parameter U(H,E,e), a function of the hypothesis being updated, the evidence producing the update, and the background information in which the update takes place. This is depicted in the upper diagram of Figure 1.

For comparison, the Bayesian conditioning scheme is represented schematically in the lower diagram of the same figure. Corresponding to the updating procedure in the belief update paradigm is the axiomatic engine of probability theory. The axiomatic engine, in turn, is "parameterized" by the full joint distribution. Inputs to the Bayesian updating procedure include the propositions of interest and outputs consist of beliefs relating to these propositions.

An important difference between the two approaches is illustrated in the figure. In the Bayesian theory, the process of updating is implicit; it is a matter of course that the belief in a given proposition changes when the conditioning propositions are modified (recall Cox's context dependency property). In contrast, the process of updating is made explicit in the update paradigm. As a consequence, the Bayesian scheme can treat hypothesis and evidence symmetrically while the update approach cannot. For example, the calculation of p(E|He) in the Bayesian approach is no different in principle then the calculation of p(H|Ee). In the update approach, however, the roles of evidence and hypothesis would have to be exchanged in order to implement the calculation of p(E|He).

In addition to the definition above, there are two fundamental properties that are ascribed to belief updates. The first property is analogous to the consistency property for absolute beliefs. It is assumed that if the arguments of a belief update are logically equivalent, then the belief updates must have the same value. That is, if $H_1 \equiv H_2$, $E_1 \equiv E_2$,

124

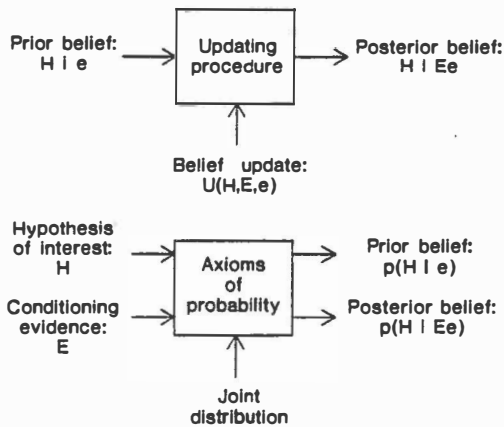

Figure 1: The belief update paradigm vs. Bayesian updating

and $e_1 \equiv e_2$, then $U(H_1,E_1,e_1) = U(H_2,E_2,e_2)$. This will be called the *consistency* property for belief updates.

The second property concerns the combination of belief updates. Consider the situation corresponding to the upper path in Figure 2. The prior belief in hypothesis H, H|e, is updated by evidence $E_1$ in the context e. Then, the posterior belief, $H|E_1e$, is updated with a second piece of evidence $E_2$. Note the third argument of the second belief update, $U(H,E_2,E_1e)$, contains $E_1$ as part of the context. The result is the belief in H given both $E_1$ and $E_2$, $H|E_1E_2e$. Alternatively, the belief in H could be updated with both pieces of evidence simultaneously as depicted in the bottom path in Figure 2. In the belief update paradigm, it is asserted that the two separate updates, $U(H,E_1,e)$ and $U(H,E_2,E_1e)$, can be combined directly to give $U(H,E_1E_2,e)$. Formally, it is assumed that there exists some function g such that

$$U(H,E_1E_2,e) = g(U(H,E_1,e), U(H,E_2,E_1e)). \qquad (8)$$

The only constraints on g is that it be continuous in both arguments and monotonically increasing in each argument when the other is held constant. This will be called the *combination* property for belief updates.

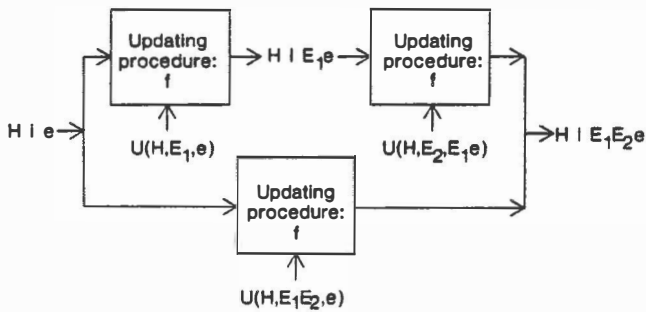

Figure 2: The combination of belief updates

## A consequence of the axioms

Although the properties above seem fairly general, they greatly restrict the quantities that may serve as belief updates. In particular, it is shown in this section that any measure which satisfies the definition of belief updates and the two properties above must also satisfy the relation:

$$h(U(H,E,e)) = i(H|Ee) - i(H|e) \qquad (9)$$

where h and i are monotonic functions. In words, a belief update $U(H,E,e)$ is simply the arithmetic difference of a posterior and prior belief, up to an arbitrary monotonic transformation. Of course, any quantity which satisfies (9) must also satisfy (7) and (8). Indeed, (9) is directly suggested by the term "update." However, in this section it is shown that (9) is a *necessary* condition for an update, a stronger result. Equation (9) will be called the *difference* property for belief updates.

Consider three items of evidence $E_1$, $E_2$, and $E_3$ for hypothesis H. Applying the combination property, (8), to $(E_1E_2)$ and $E_3$ and then to $E_1$ and $E_2$ gives:

$U(H,(E_1E_2)E_3,e)$

$= g(U(H,E_1E_2,e), U(H,E_3,E_1E_2e))$

$= g(g(U(H,E_1,e), U(H,E_2,E_1e)), U(H,E_3,E_1E_2e))$.

Equation (8) can also be used to first expand $E_1$ and $(E_2E_3)$ and then $E_2$ and $E_3$ giving:

$U(H,E_1(E_2E_3),e)$

$= g(U(H,E_1,e), U(H,E_2E_3,E_1e))$

$= g(U(H,E_1,e), g(U(H,E_2,E_1e), U(H,E_3,E_1E_2e)))$.

However, $(E_1E_2)E_3$ and $E_1(E_2E_3)$ are logically equivalent and therefore, by the consistency property, these two expressions must be equal. That is,

$$g(g(x,y), z) = g(x, g(y,z)) \qquad (10)$$

where $x = U(H,E_1,e)$, $y = U(H,E_2,E_1e)$, and $z = U(H,E_3,E_1E_2e)$. Equation (10) is called a *functional equation*. Using group theory, Aczel has shown that the most general solution to this equation is

$$h(g(x,y)) = h(x) + h(y)$$

where h is some continuous, monotonic function [14]. Therefore, it follows from the combination property, (8), that

$$h(U(H,E_1E_2,e)) \qquad (11)$$

$= h(U(H,E_1,e)) + h(U(H,E_2,E_1e)).$

Note the power of Aczel's result. It says that any continuous, monotonic function of two arguments that satisfies an associativity relation must necessarily be additive in some transformed space.

Now consider the definition of belief updates, (7). Given that the composition of two monotonic functions is another monotonic function, (7) can be rewritten as

$$H|Ee = f(h(U(H,E,e)), H|e). \qquad (12)$$

Note that the function f in (12) is not equal to the function f in (7). The same symbol is used to avoid the proliferation of unnecessary terms.

Given this new version of the definition, consider again the situation in Figure 2. The upper path in the figure corresponds to the expansion:

$H|E_2(E_1e)$

$= f(h(U(H,E_2,E_1e)), H|E_1e)$

$= f(h(U(H,E_2,E_1e)), f(h(U(H,E_1,e)), H|e))$

125

while the lower path corresponds to the expansion:

$H|(E_1E_2)e$

$= f(h(U(H,E_1E_2,e), H|e))$

$= f(h(U(H,E_1,e)) + h(U(H,E_2,E_1e)), H|e)$

In the first expansion, (12) is applied to each item of evidence separately. In the second, (12) is applied to the combined evidence $E_1$ and $E_2$ and then (11) is used to expand the update. By the consistency property, these two expansions must be equal. Therefore,

$f(x + y, z) = f(x, f(y, z))$

where $x = h(U(H,E_2,E_1e))$, $y = h(U(H,E_1,e))$, and $z = H|e$. This is another functional equation. The most general solution is

$i(f(x,y)) = x + i(y)$

where $i$ is another continuous, monotonic function [14]. Therefore, it can concluded from (7) that

$i(H|Ee) = h(U(H,E,e)) + i(H|e)$

which establishes the desired result.

## Probabilistic belief updates

In the remainder of the paper, measures of change in belief will be considered in a probabilistic context. That is, the implications of the axioms of belief updates will be explored under the assumption that each of Cox's properties are valid.

Before discussing the general case, however, it is useful to examine a particular probabilistic update. Consider the following version of Bayes' theorem for updating the probability of a hypothesis H given evidence E and the current state of information e:

$$p(H|Ee) = \frac{p(E|He)p(H|e)}{p(E|e)}. \qquad (13)$$

Note that this relationship follows directly from (5). The corresponding formula for the negation of the hypothesis, $\neg H$, is

$$p(\neg H|Ee) = \frac{p(E|\neg He)p(\neg H|e)}{p(E|e)}. \qquad (14)$$

Dividing (13) by (14) gives:

$$\frac{p(H|Ee)}{p(\neg H|Ee)} = \frac{p(E|He)}{p(E|\neg He)} \frac{p(H|e)}{p(\neg H|e)}. \qquad (15)$$

Now the odds of some event X, denoted O(X), is just

$O(X) = p(X)/p(\neg X) = p(X)/(1 - p(X))$

so that (15) can be written as

$$O(H|Ee) = \frac{p(E|He)}{p(E|\neg He)} O(H|E). \qquad (16)$$

The ratio in (16) is called a *likelihood ratio* and is written $\lambda(H,E,e)$. With this notation, (16) becomes

$$O(H|Ee) = \lambda(H,E,e) O(H|e). \qquad (17)$$

Equation (17) is called the *odds-likelihood* form of Bayes' theorem. Notice that $\lambda(H,E,e)$ and the prior odds are sufficient to determine the posterior odds. Moreover, since the odds of any event is a monotonic function of the probability of the event, it follows from (17) that the likelihood ratio $\lambda(H,E,e)$ satisfies the definition of a belief update (7).

It is also straightforward to show that $\lambda(H,E,e)$ satisfies the combination property for updates, (8). Consider two items of evidence $E_1$ and $E_2$. The likelihood ratio for the combined evidence $E_1$ and $E_2$ is

$$\lambda(H,E_1E_2,e) = \frac{p(E_1E_2|He)}{p(E_1E_2|\neg He)}.$$

Using the product rule (5), both the numerator and denominator in the above expression can be expanded giving:

$$\frac{p(E_1E_2|He)}{p(E_1E_2|\neg He)} = \frac{p(E_1|He)}{p(E_1|\neg He)} \frac{p(E_2|HE_1e)}{p(E_2|\neg HE_1e)}.$$

From the definition of $\lambda$ it follows that

$$\lambda(H,E_1E_2,e) = \lambda(H,E_1,e) \lambda(H,E_2,E_1e). \qquad (18)$$

Thus, the likelihood ratio $\lambda(H,E,e)$ satisfies the combination property, (8), where the function g is simple multiplication. Moreover, since the consistency property for updates is trivially satisfied in a probabilistic context, it follows that the likelihood ratio $\lambda$ is a legitimate belief update.

The quantity $\lambda$ has several interesting properties. For example, $\lambda$ satisfies the difference property, as it must given the previous discussion. In particular, taking the logarithm of (17) and subtracting gives

$\log[\lambda(H,E,e)] = \log[O(H|Ee)] - \log[O(H|e)].$

Another interesting property arises from assumptions of probabilistic independence. Suppose knowing $E_1$ does not influence one's belief in $E_2$ if it is known that either H or $\neg H$ is true. That is,

$p(E_2|HE_1e) = p(E_2|He)$ and $\qquad (19)$

$p(E_2|\neg HE_1e) = p(E_2|\neg He).$

Of course, this relationship is conditioned on the current state of information e. When (19) holds, it is said that $E_2$ is *conditionally independent* of $E_1$ given H and $\neg H$. With this assumption, it immediately follows from the definition of $\lambda(H,E,e)$ that

$$\lambda(H,E_2,E_1e) = \lambda(H,E_2,e). \qquad (20)$$

That is, the belief update for $E_2$ given $E_1$ does not depend on $E_1$. More generally,

$$U(H,E_2,E_1e) = U(H,E_2,e). \qquad (21)$$

Equation (21) will be called the *modularity* property for belief updates. This term is used because the above property closely resembles the informal notion of modularity associated with rule based systems [15].

Notice that the conditional independence assumption, (19), and the modularity property, (21), are both assumptions of independence but relate to different ways of thinking about the association between evidence and hypothesis. In asserting (19), one imagines that a *hypothesis* is either true or false with certainty and then contemplates the relationship between

126

two pieces of evidence for the hypothesis. In asserting (21), one imagines that a *piece of evidence* is certain and then considers how this affects the updating of the hypothesis by a second piece of evidence. From above, it is clear that these two independence conditions are closely related in a probabilistic context. In particular, when the identification $U(H,E,e) = \lambda(H,E,e)$ is made, it follows that

$$p(E_2|HE_1 e) = p(E_2|He) \quad \text{and}$$

$$p(E_2|\neg HE_1 e) = p(E_2|\neg He)$$

$$\Rightarrow U(H,E_2,E_1 e) = U(H,E_2,e). \quad (22)$$

This will be referred to as the *independence correspondence* for probabilistic belief updates.

In the remainder of this section, a general result concerning the independence correspondence will be derived. In particular, it will be shown that *any* probabilistic belief update satisfying the independence correspondence must be some monotonic transformation of $\lambda$.

To begin, consider the difference property in a probabilistic context[6]:

$$h(U(H,E,e)) = i(p(H|Ee)) - i(p(H|e)). \quad (23)$$

Because i is monotonic, (23) can be rewritten as

$$h(U(H,E,e))$$
$$= \log[j(p(H|Ee)/1-p(H|Ee))]$$
$$\quad - \log[j(p(H|e)/1-p(H|e))]$$
$$= \log[j(O(H|Ee))/j(O(H|e))] \quad (24)$$

where j is another continuous, monotonic function. Now when the conditional independence assumption, (19), is valid, it follows from Bayes' theorem that

$$\frac{O(H|E_2 E_1 e)}{O(H|E_1 e)} = \frac{O(H|E_2 e)}{O(H|e)}.$$

In addition, it follows from (24) and the modularity property that

$$j(O(H|E_2 E_1 e))/j(O(H|E_1 e))$$
$$= j(O(H|E_2 e))/j(O(H|e)).$$

Therefore, the independence correspondence implies

$$w/x = y/z \Rightarrow j(w)/j(x) = j(y)/j(z)$$

where $w = O(H|E_1 E_2 e)$, $x = O(H|E_1 e)$, $y = O(H|E_2 e)$, and $z = O(H|e)$. The most general solution is [14]:

$$j(x) = \alpha x^A$$

where A and $\alpha$ are constants. This means that

$$i(x) = A \cdot \log[x/1-x]$$

and so

$$h(U(H,E,e)) = A \cdot \log \frac{O(H|Ee)}{O(H|e)} = A \cdot \log[\lambda(H,E,e)]$$

or

$$U(H,E,e) = h^{-1}\{A \cdot \log[\lambda(H,E,e)]\}$$

which establishes the desired result.

Thus, the likelihood ratio $\lambda$ is a general belief update in the probabilistic context. The quantity $\lambda$ and monotonic transformations of it are the only measures which satisfy the axioms of belief updates in addition to the correspondence between probabilistic conditional independence and modularity.

## Conclusions

In this paper, a formal characterization of the belief update paradigm has been presented and several consequences of the characterization have been demonstrated. It is hoped that this explication will foster critical discussion and useful extensions of the approach.

## Acknowledgements

I wish to thank Eric Horvitz for help with the development of this paper. I thank Eric Horvitz, Judea Pearl, and Peter Cheeseman for insightful discussions concerning belief updates. Support for this work was provided by the Josiah Macy, Jr. Foundation, the Henry J. Kaiser Family Foundation, and the Ford Aerospace Corporation. Computing facilities were provided by the SUMEX-AIM resource under NIH grant RR-00785.

## Notes

[1] The terms "weight of evidence" [3] "measure of confirmation" [1, 5, 7] and "measure of corroboration" [12] have also been ascribed to this quantity.

[2] An exception is the formalization of belief *measurement* in the Bayesian theory [4, 10]. However, the theory does not attempt to formalize the process of belief *construction*.

[3] The components of reasoning under uncertainty and the limited scope of most methods for reasoning under uncertainty are discussed in more detail in [2].

[4] The terminology for fundamental properties of belief is introduced in [2].

[5] Actually, the function need only be strictly monotonic in the interior of its domain. For example, when P is false, PQ will also be false no matter what the value of Q|Pe. Therefore, F is not increasing in its second argument when P|e takes on the extreme value corresponding to "FALSE." This caveat applies to all functions of two arguments mentioned in this paper that are required to be monotonic.

[6] The function i should be renamed since, by Cox's result, H|e and p(H|e) are related by a monotonic transformation. As before, the same name will be retained to avoid the proliferation of notation.